\documentclass[runningheads]{llncs}
\usepackage{graphicx}
\usepackage{amsmath}
\usepackage{amssymb}
\usepackage{multirow}

\begin{document}
\title{CoMoFusion: Fast and High-quality Fusion of Infrared and Visible Image with Consistency Model}
%
\titlerunning{CoMoFusion}

\author{Zhiming Meng\inst{1}\orcidID{0009-0007-3200-7273} \and
Hui Li\inst{*1}\orcidID{0000-0003-4550-7879}\and
Zeyang Zhang\inst{1}\orcidID{0000-0003-1834-0559} \and
Zhongwei Shen\inst{2}\orcidID{0000-0002-6701-1965} \and
Yunlong Yu \inst{3}\orcidID{0000-0002-0294-2099} \and
Xiaoning Song\inst{1}\orcidID{ 0000-0002-5741-9318}\and
Xiaojun Wu\inst{1}\orcidID{0000-0002-0310-5778}
}
\authorrunning{Zhiming Meng et al.}
%
\institute{International Joint Laboratory on Artificial Intelligence of Jiangsu Province,
School of Artificial Intelligence and Computer Science, Jiangnan University,
Wuxi 214122, China \and
School of Electronic and Information Engineering, Suzhou University of Science and Technology, Suzhou, China
\and
College of Information Science and Electronic Engineering, Zhejiang University
\\
}

%
\maketitle              
\begin{abstract}
Generative models are widely utilized to model the distribution of fused images in the field of infrared and visible image fusion. 
However, current generative models based fusion methods often suffer from unstable training and slow inference speed. To tackle this problem, a novel fusion method based on consistency model is proposed, termed as CoMoFusion, which can generate the high-quality images and achieve fast image inference speed. In specific, the consistency model is used to construct multi-modal joint features in the latent space with the forward and reverse process. 
Then, the infrared and visible features extracted by the trained consistency model are fed into fusion module to generate the final fused image. 
In order to enhance the texture and salient information of fused images, a novel loss based on pixel value selection is also designed. Extensive experiments on public datasets illustrate that our
method obtains the SOTA fusion performance compared with the existing fusion methods. The source code is available at https://github.com/ZhimingMeng/CoMoFusion.
\keywords{Image fusion \and Multi-modal information \and Consistency model \and Diffusion.}
\end{abstract}

\section{Introduction}

\begin{figure}
\centering
\includegraphics[width=\linewidth]{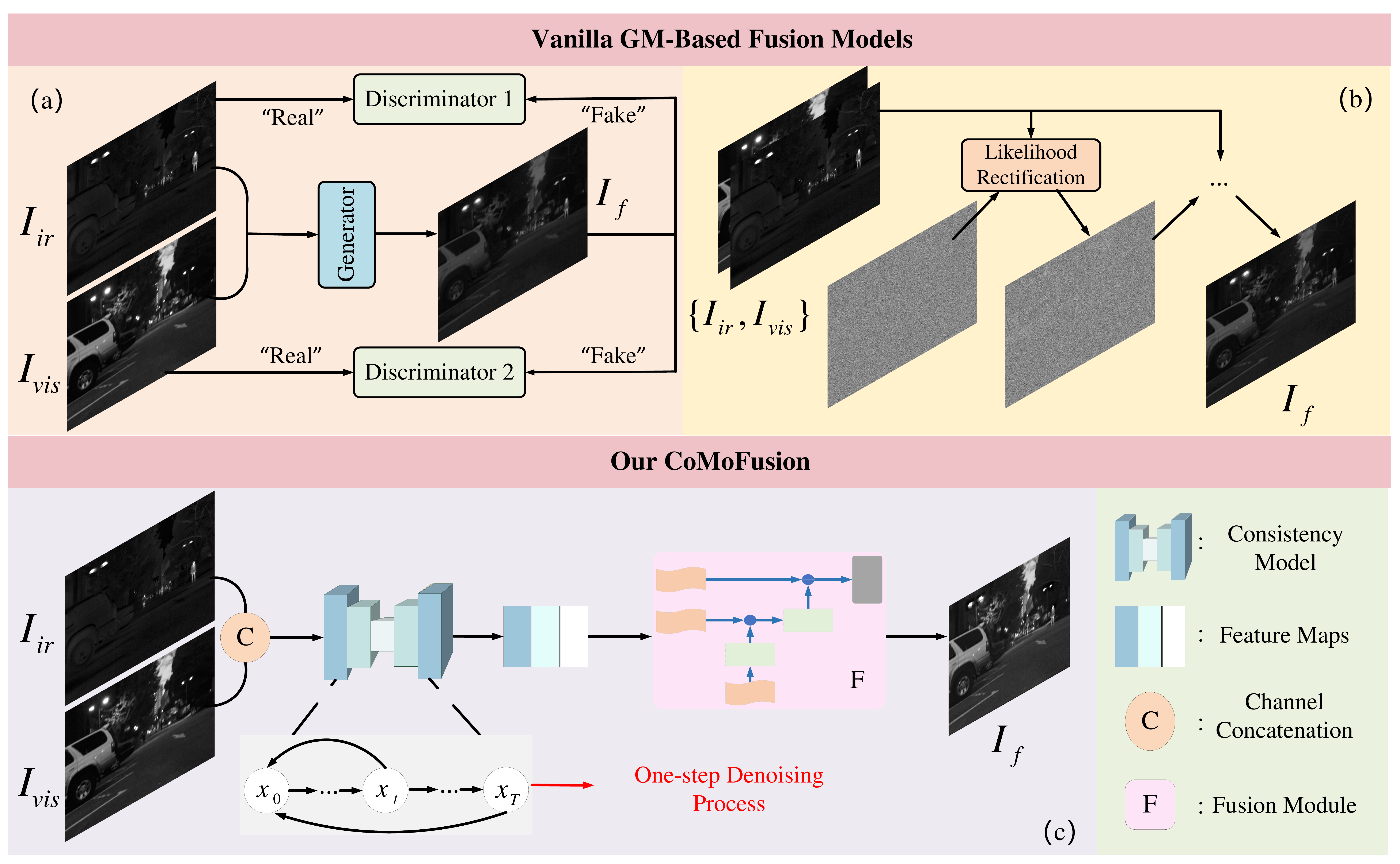}
\caption{(a) The workflow of existing GAN-based methods (GANMcC~\cite{ma2020ganmcc}). (b) The workflow of existing DDPM-based methods (DDFM~\cite{zhao2023ddfm}). (c) The workflow of our method.} \label{fig1}
\end{figure}

Due to the limitations of the optical imaging hardware equipment, the image acquired by a single sensor can only capture part of the scene information~\cite{xu2020u2fusion,zhang2021image}. 
Therefore, the image fusion technique plays an important role in computer vision field, it can acquire more comprehensive scene information from multiple source images~\cite{yue2023dif}.

The Infrared-Visible image Fusion(IVF), as an important branch of image fusion task, has played a significant role in some downstream visual tasks, such as multi-modal salient detection~\cite{wang2021dual}, object detection~\cite{bochkovskiy2020yolov4} and semantic segmentation~\cite{qin2022bibert,qin2023distribution}. Specifically, the aim of IVF is to preserve both texture information from visible images and thermal radiation information from infrared images. 
This liberalizes the working environment of the sensing system since the visible images are sensitive to illumination conditions and the infrared images are of low resolution and easy to get noised.

In the past years, researchers have widely applied generative models (GM)~\cite{goodfellow2014generative,ho2020denoising} into image fusion task.
Among them, fusion methods based on Generative Adversarial Networks (GAN)~\cite{goodfellow2014generative} are prevalent~\cite{MA201911,ma2020ganmcc}. The workflow of GAN-based fusion models, shown in Fig.~\ref{fig1} (a), often contains a generator and a discriminator. The generator creates fused image, while the discriminator determines whether the fused image has same distribution with source images. Although GAN-based methods obtain great fusion performance, they often suffer from unstable training and mode collapse, leading to the poor quality of the generated samples. In order to solve this issue, Zhao et al. introduce the denoising diffusion probabilistic model (DDPM)~\cite{ho2020denoising} into the field of IVF which achieves more stable training~\cite{zhao2023ddfm}. However, as shown in Fig.~\ref{fig1} (b), the iterative nature of diffusion model leads to a slow image generation speed, which limits the application of image fusion algorithms.

Recently, consistency model~\cite{song2023consistency} has aroused widely attention in the image generation field, which can generate high-quality images from noise-corrupted images by one-step denoising process.
Compared to GAN~\cite{goodfellow2014generative} and DDPM~\cite{ho2020denoising}, consistency model exhibits a more stable training process and faster image generation speed, respectively. 


Therefore, the consistency model is firstly applied into image fusion task and a novel infrared and visible image fusion method is proposed, named CoMoFusion, as shown in Fig.~\ref{fig1} (c). 
First of all, the infrared and visible images are concatenated on channel dimension and fed into consistency model to construct multi-modal joint features in the latent space. With the effective training process, consistency model can extract more robust features from source images. 
Secondly, the source image features extracted by the trained consistency model are fed into the proposed fusion module to generate fused image.
Moreover, in order to enhance the texture and salient features of fused image, a novel pixel-value-selection ($L_{pvs}$) based loss function is designed.
With $L_{pvs} $, the proposed fusion network can adaptively select useful parts from source images to enhance the quality of fused images. 

The main contributions of this paper can be summarized as follows:
\begin{itemize}
\item[$\bullet$] An novel fusion network based on consistency model is proposed which has a stable training process and achieves fast running time in testing phase. 
\item[$\bullet$] We design a new loss function based on pixel value selection which constrains fused results to preserve more complementary information from source images.
\item[$\bullet$] Extensive experiments on public datasets demonstrate that the proposed method achieves better fusion performance on subjective and objective evaluation.

\end{itemize}

\section{Related Work}
In this section, we give a brief introduction to deep learning based IVF models and consistency model.

\subsection{Deep Learning based IVF Methods}
With the popularity of deep learning in the computer vision community, it has been widely applied to the field of image fusion, such as autoencoder (AE) based methods~\cite{li2018densefuse}, end-to-end fusion methods~\cite{xu2020u2fusion,tang2022piafusion}, and GM based methods~\cite{zhu2017fusion,ma2020ganmcc,zhao2023ddfm}. 

For AE based methods, DenseFuse~\cite{li2018densefuse} is a typical case. In this work, the encoding network is composed of dense blocks, fusion layer and convolutional layers, while the decoding network generates fusion results. However, its fusion rule needs to be set manually, which may not be suitable for complex scenarios. To avoid manual design of fusion rules, Li et al. improved DenseFuse by designing an end-to-end fusion model~\cite{li2021rfn}, which introduces a learnable fusion strategy based on convolutional layers.

\begin{figure}
\includegraphics[width=\textwidth]{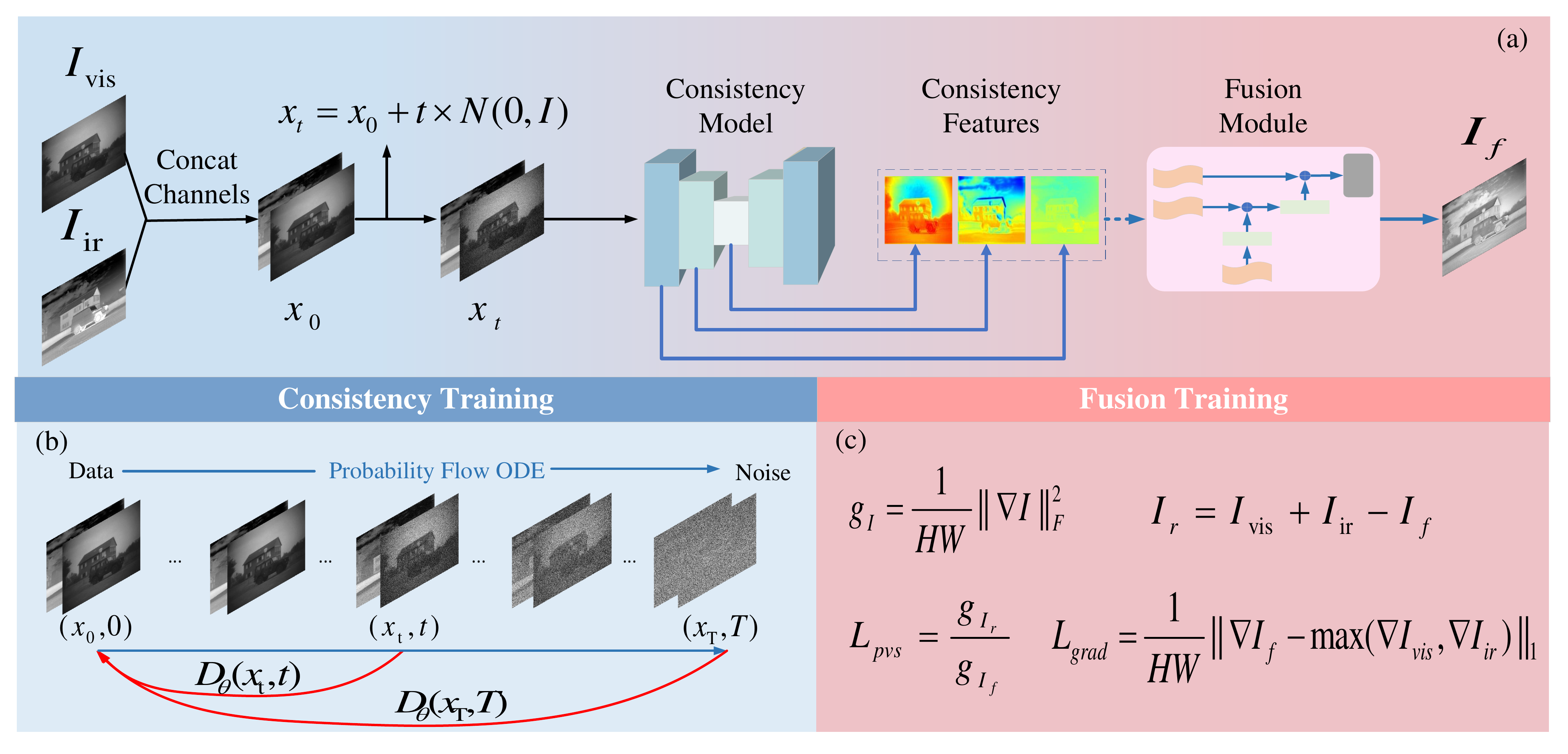}
\caption{The framework of CoMoFusion. (a)Two training stages: consistency training, fusion training.  (b)The forward and reverse process of consistency model training. (c)The loss function of fusion training.} \label{fig2}
\end{figure}

In end-to-end fusion methods, Tang et al. proposed a progressive infrared and visible image fusion network based on 
illumination aware~\cite{tang2022piafusion}. 
 To ensure that fused images retain more information from source images, Xu et al.~\cite{xu2020u2fusion} used the feature maps of pre-trained VGG-16 network~\cite{simonyan2014very} to measure the information of source images quantitatively for adaptive allocation of loss function weights. 
 
 For GM based methods, Ma et al. firstly applied GAN to IVF named FusionGAN~\cite{ma2019fusiongan}. In order to mitigate the issues of gradient explosion and vanishing gradients during the training process, as well as to enhance the training effectiveness of the generator, GANMcC~\cite{ma2020ganmcc} was proposed. In ~\cite{zhao2023ddfm}, Zhao et al. first applied diffusion models to IVF, transforming IVF into a conditional generation problem under the DDPM sampling framework. However, the above-mentioned methods based on generative models are either unstable during training phase or have a slow image generation speed, limiting the practical application of image fusion algorithms in real life.


\subsection{Consistency Model}

Diffusion models which are also known as score-based generative models, have been successfully applied into multiply fields, including image generation~\cite{rombach2022high}, image inpainting~\cite{lugmayr2022repaint} and audio synthesis~\cite{chen2020wavegrad}. However, compared to single-step generative models~\cite{goodfellow2014generative}, the iterative generation procedure of diffusion models typically requires
10–2000 times iterative computation for sample generation which causes slow inference and also hard to utilize in real-time conditions~\cite{song2020improved}.

In order to solve this issue, Yong et al. proposed an one-step denoising model, named consistency model~\cite{song2023consistency}, which is based on the probability flow (PF) ordinary differential equation (ODE)~\cite{song2021scorebased} in continuous-time diffusion models. In the forward process, the original data is gradually disturbed in several timesteps by adding Gaussian noise. In the reverse process, instead of recovering the original data by predicting the added noise at each step in the forward process, consistency model establishes the relationship mapping between the original data and the noise. 

Thanks to the exceptional performance and fast sampling speed of consistency model in generation tasks, in this paper, the consistency model is firstly introduced into image fusion task which can extract more powerful deep features and achieve state-of-the-art fusion performance.

\section{Method}
In this section, we introduce our infrared and visible image fusion framework based on consistency model in detail. 

As demonstrated in Fig.~\ref{fig2} (a), firstly, the infrared and visible image pairs concatenated on channel dimension are fed into consistency model to construct multi-modal joint features. Then, we extract features that encompass both infrared and visible information from consistency model into fusion module to generate fused images with the guidance of pixel value selection loss $L_{pvs} $ and gradient loss $L_{grad} $. We describe the above process in detail in the following subsections.

\subsection{Construct Multi-modal Joint Features}
 Given a pair of registered visible image $I_{vis} \in \mathbb{R} ^{H\times W\times1} $ and infrared image $I_{ir} \in \mathbb{R} ^{H \times W\times1} $, we concatenate them on channel dimension to form multi-modal input $x_{0} \in \mathbb{R} ^{H\times W \times 2} $, which represents multi-modal data without any noises. Assuming that the distribution of $x_{0}$ is $p_{{data}}({x})$.

\subsubsection {Forward Process.}As seen in Fig.~\ref{fig2} (b), the forward process of consistency model in continuous-time can be described by the Probability Flow ODE ~\cite{song2021scorebased} as follows:
\begin{equation}
\mathrm{d}x_{t}=\sqrt{2t} \mathrm{d}w_{t}
\end{equation}
where  $t\in [\epsilon,T]$, $T$ is a fixed constant, $x_{t}$ denotes the state of $x_{0}$ at $t$ and 
$w_{t}$ is the standard wiener process.

In practice, we follow EDM\footnote{Elucidating the Design Space of Diffusion-Based
Generative Models (EDM)~\cite{karras2022elucidating}.} to discretize the time horizon $[\epsilon,T]$ into $N-1$ sub-intervals, with the boundaries $t_{1}=\epsilon<t_{2} < \cdots < t_{N} = T$, the adding-noise process of consistency model in a discrete form can be given as follows:
\begin{equation}
x_{t_{i}}=x_{0}+t_{i}z 
\quad t_i=(\epsilon^{1/\rho}+\frac{i-1}{N-1} (T^{1/\rho}-\epsilon^{1/\rho}))^\rho\
\label{add_noise}
\end{equation}
where ${z}$ represents Gaussian noise following ${z}\sim \mathcal{N}({0}, {I})$ and the addition of noise intensifies with the growth of $t_i$. Notably, we set $\rho=7$, $\epsilon=0.002$, $T=80$ and $N=40$.

\subsubsection {Reverse Process.}In the reverse process, consistency function is constructed to make any point from the same Probability Flow ODE trajectory be mapped to the same initial point, which can be represented by
\begin{equation}
D_{{\theta}}(x_{t},t)=c_{skip}(t)x_{t}+c_{out}(t)F_{{\theta}}(x_{t},t)    
\end{equation}
where $F_{{\theta}}(x_{t},t)$ is the output of consistency model, $c_{skip}(t)$ modulates the skip connection, $c_{out}(t)$ scales the magnitudes of $F_{{\theta}}(x_{t},t)$. 

In order to ensure  $c_{skip}(t)$ and $c_{out}(t)$ are differentiable and $D_{{\theta}}(x_{\epsilon},\epsilon)=x_{\epsilon}$, they are formulated as follows,

\begin{equation}
c_{\mathrm{skip}}\left(t\right)=\frac{\sigma_{\mathrm{data}}^{2}}{(t-\epsilon)^{2}+\sigma_{\mathrm{data}}^{2}},\quad c_{\mathrm{out}}\left(t\right)=\frac{\sigma_{\mathrm{data}}\left(t-\epsilon\right)}{\sqrt{\sigma_{\mathrm{data}}^{2}+t^{2}}}
\end{equation}
where $\sigma_{\mathrm{data}} $ denotes the standard deviation of  $p_{{data}}({x})$.

\subsubsection {Loss Function of Consistency Model Training Phase.}
First, we sample $i$ from the uniform distribution $\mathcal{U}(1,N-1)$, $x_{t_{i}}$ and $x_{t_{i} + 1}$ can be calculated by Equation (\ref{add_noise}). In order to stabilize the training process and improve the final performance, the exponential moving average (EMA) is applied. Then, the loss of consistency training can be given as follows,
\begin{equation}
\mathbb{E}[\lambda(t_i)d({D}_{{\theta}}({x_{t_{i+1}}},t_{i+1}),{D}_{{\theta}^{-}}({x_{t_{i}}},t_i))]
\end{equation}
where $\lambda(t_i)$ denotes the weight corresponding to different noise level $t_i$, ${\theta}^{-}$ is a running average of the past values of ${\theta}$ and $d(\cdot)$ is  the Learned Perceptual Image Patch Similarity (LPIPS) ~\cite{zhang2018unreasonable}.

\begin{figure}
\vspace{-0.8cm}
\includegraphics[width=\textwidth]{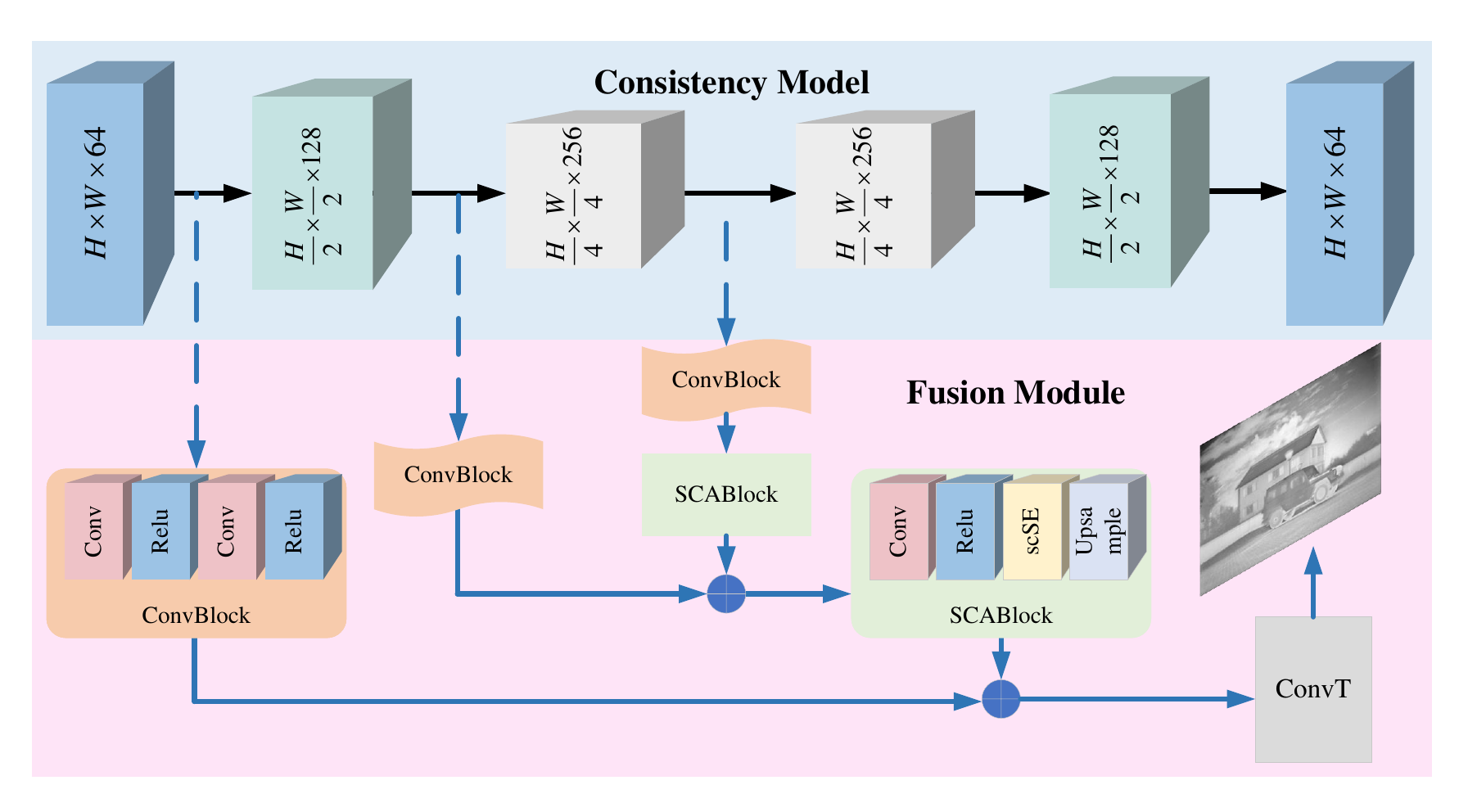}
\caption{The structure of consistency model and fusion module.} \label{fig3}
\vspace{-0.8cm}
\end{figure}

\subsection{Image Fusion with Consistency Features}
After constructing multi-modal joint features of the infrared and visible images with consistency model, we extract consistency features by input $(x_{\epsilon},\epsilon)$ into consistency model. $\epsilon$ is a tiny value, we assume that $x_{\epsilon}$ is equal to
$x_{0}$. Afterwards, the consistency features are fed into fusion module for training with 
two loss functions ($L_{pvs} $ and $L_{grad}$).

\subsubsection {Multi-modal Joint Features.} 
As depicted in Fig.~\ref{fig3}, the encoder of consistency model has three convolutional layers, and the size of output feature maps are $(H\times W)$, $(H/2\times W/2)$, $(H/4\times W/4)$. The decoder of consistency model also has three convolutional layers and the size of feature maps is opposite. Compared with the decoder, the encoder usually exhibits a stronger feature representation ability. Therefore, we utilize the features extracted by encoder for fusion. In the ablation studies, we will compare the features of encoder and decoder for fusion to provide further evidence.

\subsubsection {Fusion Module.}
Fusion Module is composed of three ConvBlocks, two SCABlocks and one convolutional layer (ConvT), as shown in Fig.~\ref{fig3}. First, three ConvBlocks are used to process the multi-scale consistency features with convolutional layer ($3\times 3$ kernel), ReLU activation functions. After that, SCABlocks consist of $3\times 3$ convolutional kernel with padding, ReLU activation functions, Concurrent Spatial and Channel Squeeze and Channel Excitation (scSE)~\cite{roy2018concurrent} and upsampling. With scSE, SCABlocks can recalibrate features along
channel and space to enhance meaningful information. 

Finally, a convolutional layer (ConvT) is adopted to generate fused images $I_{f}$ with a $3\times 3$ convolutional kernel and a Tanh activation function. 

\subsubsection {Loss Function of Fusion Module Training Phase.}
We decompose the source images into fused image $I_{f}$ and redundant images $I_{r}$. $I_{f}$ can be obtained from fusion module, $I_{r}$ is represented as follows,
\begin{equation}
I_{r}= I_{vis}+I_{ir}-I_{f}
\end{equation}

For IVF, $I_{f}$ should preserve more information including salient and texture information from source images, $I_{r}$ is opposite. Inspired by~\cite{xu2020u2fusion}, we measure the salient and texture information of the image $I$ as follows,
\begin{equation}
g_{I}=\frac{1}{H W}\|\nabla I\|_{F}^{2}
\end{equation}
where $\|\cdot\|_{F}$ denotes the Frobenius norm, $\nabla$ is the sobel operator. In order to keep $I_{f}$ more informative and $I_{r}$ less informative, the pixel value selection loss ($L_{pvs}$) is designed and calculated as follows,
\begin{equation}
L_{pvs}=\frac{g_{I_{r}}}{g_{I_{f}}}
\end{equation}

Under the constraint of $L_{pvs} $, fused image can adaptively select pixel values to retain more useful parts from the source images.
Meanwhile, gradient loss $L_{grad}$ is devised to restrain fused image retaining vital details
from the source images, which is represented as follows,
\begin{equation}
L_{grad}=\frac1{HW}\|\nabla I_f-\max(\nabla I_{vis},\nabla I_{ir})\|_1
\end{equation}
where $\|\cdot\|_1$ denotes the $l_1$ norm. To sum up, the loss function of the proposed CoMoFusion can be defined as follows,
\begin{equation}
L_{f}=L_{pvs}+\lambda L_{grad}
\end{equation}
where $\lambda$ is a weight parameter to control the trade-off. Since the magnitudes of \( L_{pvs} \) and \( L_{grad} \) are on the same order, the \( \lambda \) is set to 1 in our experiments.

\section{Experiments}
In this section, we elaborate the implementation and configuration of our networks for IVF in detail. The experiments show the fusion performance of our model and the rationality of network structures.
\subsection{Setup}
\subsubsection {Datasets and Metrics.}Our proposed model is trained on the KAIST dataset~\cite{hwang2015multispectral} (95328 pairs). TNO (42 pairs)~\cite{toet2014} and MSRS (361pairs)~\cite{tang2022piafusion} are employed as test datasets. Note that our model is not fine-tuned on the TNO and MSRS. In order to measure the performance of fusion results, six metrics\footnote{For the details of those metrics please refer to~\cite{ma2019infrared}.} are applied, including entropy (EN), spatial frequency (SF), average gradient (AG),  standard deviation (SD), Qabf and structural similarity index measure (SSIM). The higher scores of the metrics indicates the better performance of fusion results.
\begin{figure}[t]
\includegraphics[width=\textwidth]{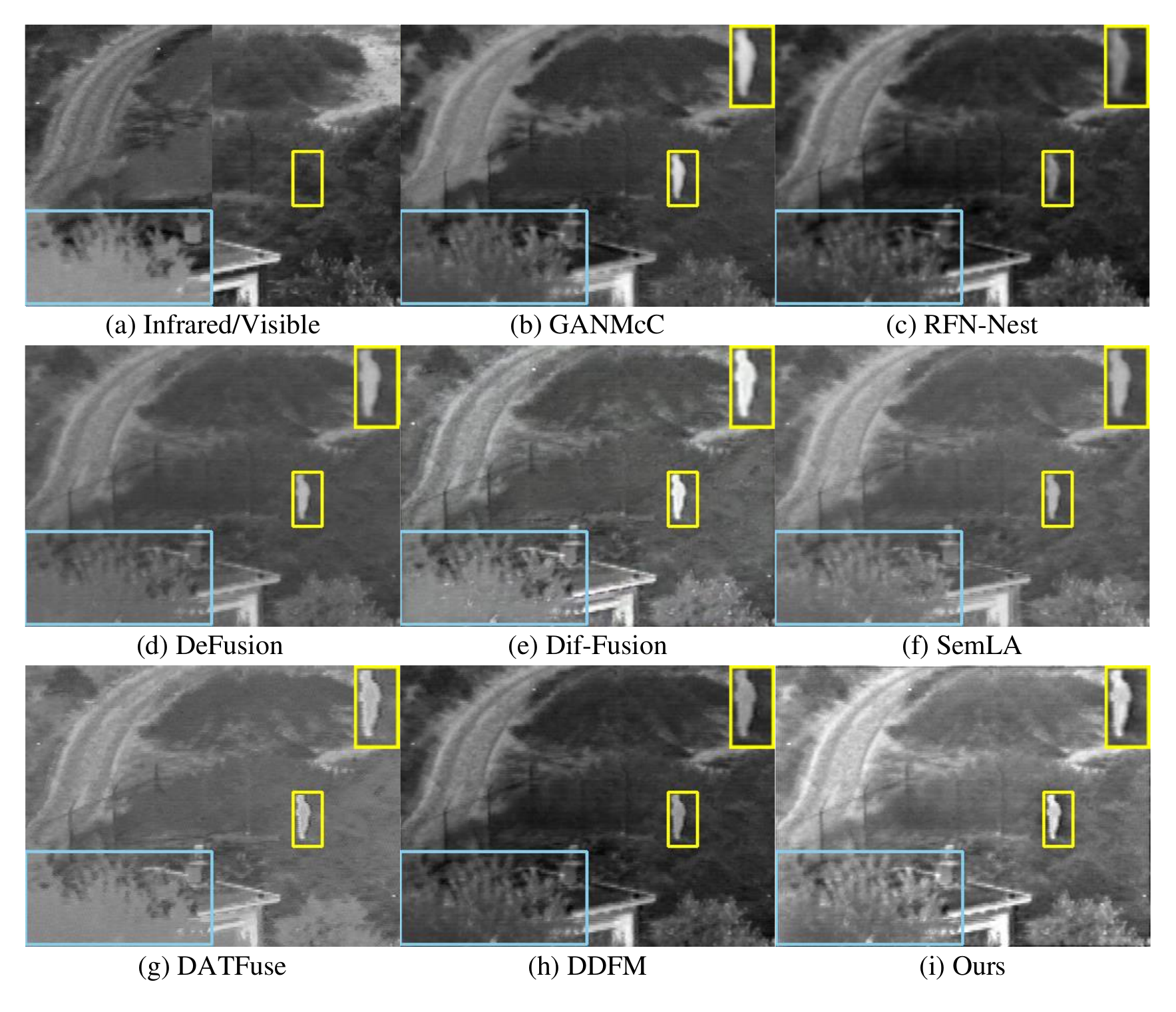}
\caption{Qualitative comparison of the image “35” in the TNO dataset.} \label{fig4}
\end{figure}

\begin{figure}[t]
\includegraphics[width=\textwidth]{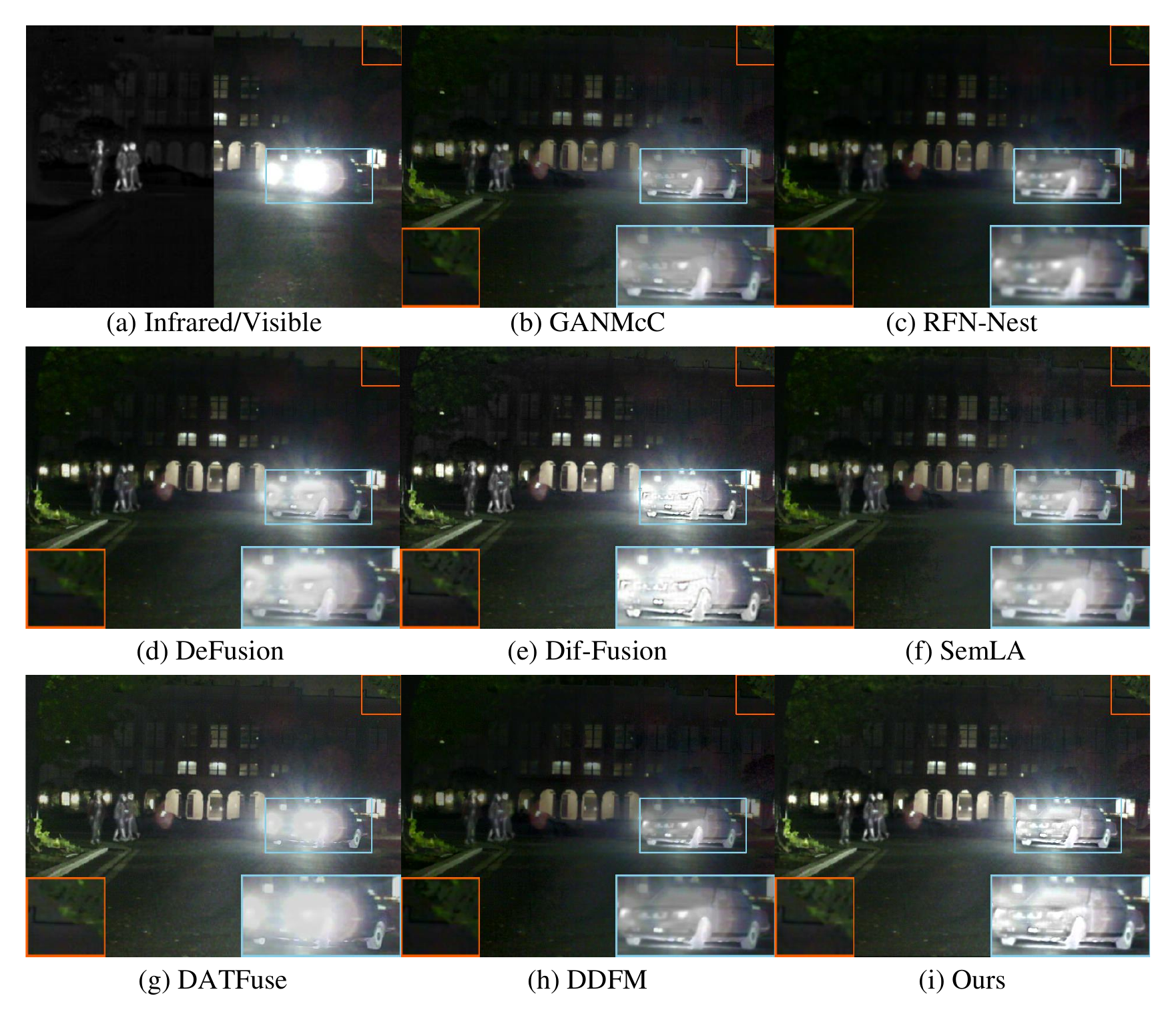}
\caption{Qualitative comparison of the image “00838N” in the MSRS dataset.} \label{fig5}
\end{figure}

\subsubsection {Implementation Details.} All experiments are conducted on the NVIDIA GeForce RTX 4090 using PyTorch as our programming environment. In the consistency training process, we adopt the training settings in ~\cite{song2023consistency}. Moreover, the training images are converted to gray scale and randomly cropped to $160 \times 160$, the batch size is set to 15. When training the fusion module, we set the batch size to 15, the learning rate and epoch are set to $1 \times 10^{-4}$ and 2, respectively.

\subsection{Comparison with Existing Methods}
Seven IVF methods are chosen to conduct the comparison experiments, including one AE based model (RFN-Nest~\cite{li2021rfn}), three end-to-end methods (DeFusion~\cite{liang2022fusion}, SemLA~\cite{XIE2023101835} and DATFuse~\cite{tang2023datfuse} ), and three GM based methods (GANMcC~\cite{xu2020u2fusion}, Dif-Fusion~\cite{yue2023dif} and DDFM~\cite{zhao2023ddfm}).  

\subsubsection {Qualitative Comparison.} As shown in Fig.~\ref{fig4} and Fig.~\ref{fig5}, our method highlights salient regions while preserving texture information. For example, in  Fig.~\ref{fig4}, only our method, DDFM and RFN-Nest perform well in maintaining texture details from the visible image (blue box). However, DDFM and RFN-Nest exhibits poor performance in retaining thermal information from the infrared image (yellow box). Meanwhile, in the regions marked by orange and blue boxes in Fig.~\ref{fig5}, our method
shows clearer targets not only in bright-light condition (the car), but also in dim-light condition (the foliage).

\begin{table}
 \centering
\caption{Quantitative comparison on the TNO dataset. The \textbf{bold} and \underline{underlined} part show the best and second-best values, respectively.}\label{tab1}
\setlength{\tabcolsep}{2.5mm}
{
\begin{tabular}{cccccccc}
\hline
 Method&     Year    &  EN$\uparrow$ & SF$\uparrow$ & AG$\uparrow$ & SD$\uparrow$  & Qabf$\uparrow$ &SSIM$\uparrow$ \\
\hline
GANMcC~\cite{ma2020ganmcc}& 2020 & 6.736& 6.6161 & 2.544 &33.437 &0.281&0.422\\
RFN-Nest~\cite{li2021rfn} & 2021 & \underline{6.963}&5.874 & 2.669 &36.897 &0.335&0.398\\
DeFusion~\cite{liang2022fusion} & 2022 & 6.573& 6.375 & 2.607 &31.253 &0.376&0.458\\
Dif-Fusion~\cite{yue2023dif} & 2023 & 6.925&  \underline{10.673} & \underline{4.26} &\underline{38.873} &0.467&0.434\\
SemLA~\cite{XIE2023101835}& 2023 & 6.655& 9.169 & 3.253 &32.614 &0.368&0.416\\
DATFuse~\cite{tang2023datfuse} & 2023 & 6.453& 9.606 & 3.56 &27.576 &\textbf{0.5}&0.469\\
DDFM~\cite{zhao2023ddfm} & 2023 & 6.849& 8.528 & 3.372 &34.26 &0.434&\textbf{0.503}\\
CoMoFusion & ours & \textbf{7.081}&  \textbf{14.093} & \textbf{5.069} &\textbf{39.844}&\underline{0.482} &\underline{0.491}\\
\hline
\end{tabular}
}

\end{table}

\begin{table}

 \centering
\caption{Quantitative comparison on the MSRS dataset. The \textbf{bold} and \underline{underlined} part show the best and second-best values, respectively.}\label{tab2}
\setlength{\tabcolsep}{2.5mm}
{
\begin{tabular}{cccccccc}
\hline
 Method &     Year    &  EN$\uparrow$ & SF$\uparrow$ & AG$\uparrow$ & SD$\uparrow$  & Qabf$\uparrow$ &SSIM$\uparrow$ \\
\hline
GANMcC ~\cite{ma2020ganmcc} & 2020 & 6.12& 5.664 & 2.006 &26.052 &0.302&0.393\\
RFN-Nest ~\cite{li2021rfn} & 2021 & 6.196& 6.167 & 2.122 &29.088 &0.388&0.375\\
DeFusion ~\cite{liang2022fusion}& 2022 & 6.459& 8.606 & 2.781 &37.913 &0.53&\underline{0.458}\\
Dif-Fusion ~\cite{yue2023dif} & 2023 & \underline{6.661}& 8.312 & \underline{3.89} &\textbf{41.902} &0.583&0.448\\
SemLA ~\cite{XIE2023101835}& 2023 & 6.217& 8.312 & 2.687 &29.374 &0.431&0.395\\
DATFuse~\cite{tang2023datfuse} & 2023 & 6.48& \underline{10.927} & 3.574 &36.476 &\textbf{0.64}&0.452\\
DDFM~\cite{zhao2023ddfm} & 2023 & 6.175& 7.388 & 2.522 &28.925 &0.474&0.453\\
CoMoFusion & ours & \textbf{6.712}&  \textbf{11.847} & \textbf{3.906} &\underline{41.533} &\underline{0.622}&\textbf{0.48}\\
\hline
\end{tabular}
}

\end{table}

\subsubsection {Quantitative Comparison.}Table~\ref{tab1} and ~\ref{tab2} demonstrate the quantitative comparison of various methods. For each metric, the best and
the second best methods are marked in bold and underlined. Our proposed method achieves higher values than other methods in EN, SF and AG which means our fused image has more information and retains richer texture details. Meanwhile, our method also demonstrates competitive performance in SD, Qabf and SSIM which are consistent with human visual perception.

\subsubsection {Inference Time Comparison.}We evaluate the average inference time of all methods on the TNO and MSRS dataset using the NVIDIA GeForce RTX 4090, and the results are presented in Table~\ref{tab4}. Among the methods compared, CoMoFusion achieves the second and first rankings on the two datasets, respectively. Meanwhile, CoMoFusion demonstrates the fastest image generation speed in the GM based methods which is friendly to the application of downstream real-time tasks.

 
\begin{table}

 \centering
\caption{Comparison of the average inference time of one image on the two datasets. The \textbf{blod} and \underline{underlined} part show the best and second-best values, respectively.}\label{tab4}
\setlength{\tabcolsep}{6mm}
{
\begin{tabular}{cccc}
\hline
  Type &Method  &  TNO(s)$\downarrow$ & MSRS(s)$\downarrow$\\
\hline

 \multirow{4}{*}{non-GM-based}&RFN-Nest~\cite{li2021rfn}&0.1314 &0.0910 \\
 &DeFusion ~\cite{liang2022fusion} &0.098 &0.0498 \\
 &SemLA ~\cite{XIE2023101835}&3.1769 &3.3145\\
  &DATFuse ~\cite{tang2023datfuse}&\textbf{0.0145} & \underline{0.0098}\\
\hline
 \multirow{3}{*}{GM-based}& GANMcC ~\cite{ma2020ganmcc}&0.0923& 0.0567\\
 &Dif-Fusion~\cite{yue2023dif} &1.7395  & 0.8333\\
 &DDFM ~\cite{zhao2023ddfm}&35.1243 &33.9328\\
\hline
 ours &CoMoFusion &\underline{0.0221}&\textbf{0.0045} \\
\hline
\end{tabular}
}

\end{table}

\begin{table}
 \centering
\caption{The objective results of ablation study on the MSRS dataset. \textbf{Bold} indicates the best.}\label{tab3}
\setlength{\tabcolsep}{4.2mm}
{
\begin{tabular}{cccccc}
\hline
 &     Configs    &      EN$\uparrow$  & SD$\uparrow$&Qabf$\uparrow$&SSIM$\uparrow$ \\
\hline
I & w/o $L_{pvs}$ & \textbf{6.937}& 39.871 & 0.598&-0.18\\
II & w/o $L_{grad}$ & 6.69& 39.928 & 0.592&0.441 \\
III & EF $\rightarrow$ DF & 6.696& 41.216 &0.62& 0.472 \\
IV & w/o CM & 6.7& 40.879 & 0.616&0.466 \\
\hline
 & ours & 6.712&  \textbf{41.533} & \textbf{0.622}& \textbf{0.48} \\

\hline
\end{tabular}
}
\end{table}
\begin{figure}
\vspace{-0.6cm}
\includegraphics[width=\textwidth]{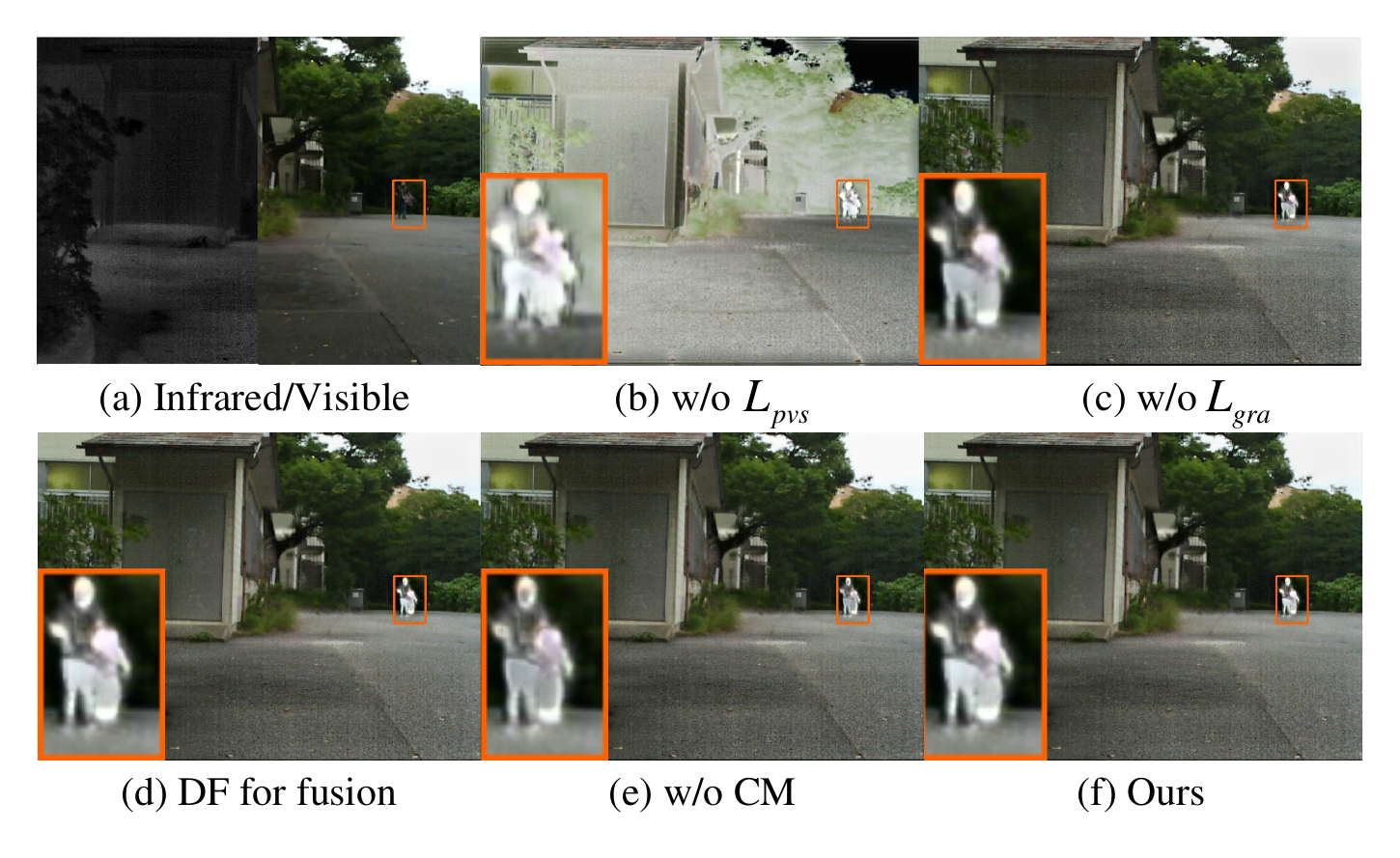}
\caption{The visualization results of ablation study with different settings.} \label{fig6}
\vspace{-0.4cm}
\end{figure}

\subsection{Alation Study}
We conduct ablation experiments to verify the rationality of module design. All experiments are trained on KAIST and tested on MSRS. EN, SD, Qabf, SSIM
are selected to validate the effective of fusion results quantitatively. The Qualitative and Quantitative results are shown in Fig.~\ref{fig6} and Table~\ref{tab3}.
\subsubsection {Loss Function.} In Exp. I, we remove $L_{pvs}$ in the loss function. As shown in Fig.~\ref{fig6} (b) , the fusion result hardly retails luminance information
  from the source images, leading to the poor performace in SSIM.
After removing  $L_{grad}$ in Exp. II, the clarity of the pedestrian is diminished in Fig.~\ref{fig6} (c). Meanwhile, there is a decrease in the metrics as well.

\subsubsection {EF for fusion or DF for fusion.} We replace the encoding features (EF) of consistency model with the the decoding features (DF) of consistency model for fusion in Exp. III. While the visualized result is close to ours in Fig.~\ref{fig6} (d), the four inferior
 metrics indicate that EF exhibits a stronger feature representation ability comparing to DF, which is more conducive to image fusion.

\subsubsection {Consistency Model.} Finally, in order to assess the effectiveness of the consistency model (CM) objectively, we introduce an autoencoder network based on a UNet-style architecture as a substitute for consistency model (CM) in Exp. IV. Aftering removing consistency model, the qualitative and quantitative results are not satisfactory. It 
is proved that consistency model can extract more robust features from source images because of its consistency training.

\section{Conclusion}
In this paper, an infrared and visible image fusion method based on consistency model (CoMoFusion) is proposed to alleviate the drawbacks of existing GM-based fusion methods. With consistency model, CoMoFusion can extract more robust features from sources images for fusion and achieve fast image inference speed. Moreover, a novel loss function based on pixel value selection is proposed to enhance the texture and salient features of fused image. Extensive experiments on public datasets show that the proposed method outperforms seven other state-of-the-art methods in fusion performance, as evaluated objectively and subjectively.


%
%
%
\bibliographystyle{splncs04}
\bibliography{edbd}

\begin{thebibliography}{10}
\providecommand{\url}[1]{\texttt{#1}}
\providecommand{\urlprefix}{URL }
\providecommand{\doi}[1]{https://doi.org/#1}

\bibitem{bochkovskiy2020yolov4}
Bochkovskiy, A., Wang, C.Y., Liao, H.Y.M.: Yolov4: Optimal speed and accuracy of object detection. arXiv preprint arXiv:2004.10934  (2020)

\bibitem{chen2020wavegrad}
Chen, N., Zhang, Y., Zen, H., Weiss, R.J., Norouzi, M., Chan, W.: Wavegrad: Estimating gradients for waveform generation. arXiv preprint arXiv:2009.00713  (2020)

\bibitem{goodfellow2014generative}
Goodfellow, I., Pouget-Abadie, J., Mirza, M., Xu, B., Warde-Farley, D., Ozair, S., Courville, A., Bengio, Y.: Generative adversarial nets. Advances in neural information processing systems  \textbf{27} (2014)

\bibitem{ho2020denoising}
Ho, J., Jain, A., Abbeel, P.: Denoising diffusion probabilistic models. Advances in neural information processing systems  \textbf{33},  6840--6851 (2020)

\bibitem{hwang2015multispectral}
Hwang, S., Park, J., Kim, N., Choi, Y., So~Kweon, I.: Multispectral pedestrian detection: Benchmark dataset and baseline. In: Proceedings of the IEEE conference on computer vision and pattern recognition. pp. 1037--1045 (2015)

\bibitem{karras2022elucidating}
Karras, T., Aittala, M., Aila, T., Laine, S.: Elucidating the design space of diffusion-based generative models. Advances in Neural Information Processing Systems  \textbf{35},  26565--26577 (2022)

\bibitem{li2018densefuse}
Li, H., Wu, X.J.: Densefuse: A fusion approach to infrared and visible images. IEEE Transactions on Image Processing  \textbf{28}(5),  2614--2623 (2018)

\bibitem{li2021rfn}
Li, H., Wu, X.J., Kittler, J.: Rfn-nest: An end-to-end residual fusion network for infrared and visible images. Information Fusion  \textbf{73},  72--86 (2021)

\bibitem{liang2022fusion}
Liang, P., Jiang, J., Liu, X., Ma, J.: Fusion from decomposition: A self-supervised decomposition approach for image fusion. In: European Conference on Computer Vision. pp. 719--735. Springer (2022)

\bibitem{lugmayr2022repaint}
Lugmayr, A., Danelljan, M., Romero, A., Yu, F., Timofte, R., Van~Gool, L.: Repaint: Inpainting using denoising diffusion probabilistic models. In: Proceedings of the IEEE/CVF conference on computer vision and pattern recognition. pp. 11461--11471 (2022)

\bibitem{ma2019infrared}
Ma, J., Ma, Y., Li, C.: Infrared and visible image fusion methods and applications: A survey. Information fusion  \textbf{45},  153--178 (2019)

\bibitem{MA201911}
Ma, J., Yu, W., Liang, P., Li, C., Jiang, J.: Fusiongan: A generative adversarial network for infrared and visible image fusion. Information Fusion  \textbf{48},  11--26 (2019). \doi{https://doi.org/10.1016/j.inffus.2018.09.004}, \url{https://www.sciencedirect.com/science/article/pii/S1566253518301143}

\bibitem{ma2019fusiongan}
Ma, J., Yu, W., Liang, P., Li, C., Jiang, J.: Fusiongan: A generative adversarial network for infrared and visible image fusion. Information fusion  \textbf{48},  11--26 (2019)

\bibitem{ma2020ganmcc}
Ma, J., Zhang, H., Shao, Z., Liang, P., Xu, H.: Ganmcc: A generative adversarial network with multiclassification constraints for infrared and visible image fusion. IEEE Transactions on Instrumentation and Measurement  \textbf{70},  1--14 (2020)

\bibitem{qin2022bibert}
Qin, H., Ding, Y., Zhang, M., Yan, Q., Liu, A., Dang, Q., Liu, Z., Liu, X.: Bibert: Accurate fully binarized bert. arXiv preprint arXiv:2203.06390  (2022)

\bibitem{qin2023distribution}
Qin, H., Zhang, X., Gong, R., Ding, Y., Xu, Y., Liu, X.: Distribution-sensitive information retention for accurate binary neural network. International Journal of Computer Vision  \textbf{131}(1),  26--47 (2023)

\bibitem{rombach2022high}
Rombach, R., Blattmann, A., Lorenz, D., Esser, P., Ommer, B.: High-resolution image synthesis with latent diffusion models. In: Proceedings of the IEEE/CVF conference on computer vision and pattern recognition. pp. 10684--10695 (2022)

\bibitem{roy2018concurrent}
Roy, A.G., Navab, N., Wachinger, C.: Concurrent spatial and channel ‘squeeze \& excitation’in fully convolutional networks. In: Medical Image Computing and Computer Assisted Intervention--MICCAI 2018: 21st International Conference, Granada, Spain, September 16-20, 2018, Proceedings, Part I. pp. 421--429. Springer (2018)

\bibitem{simonyan2014very}
Simonyan, K., Zisserman, A.: Very deep convolutional networks for large-scale image recognition. arXiv preprint arXiv:1409.1556  (2014)

\bibitem{song2023consistency}
Song, Y., Dhariwal, P., Chen, M., Sutskever, I.: Consistency models. arXiv preprint arXiv:2303.01469  (2023)

\bibitem{song2020improved}
Song, Y., Ermon, S.: Improved techniques for training score-based generative models. Advances in neural information processing systems  \textbf{33},  12438--12448 (2020)

\bibitem{song2021scorebased}
Song, Y., Sohl-Dickstein, J., Kingma, D.P., Kumar, A., Ermon, S., Poole, B.: Score-based generative modeling through stochastic differential equations. In: International Conference on Learning Representations (2021), \url{https://openreview.net/forum?id=PxTIG12RRHS}

\bibitem{tang2022piafusion}
Tang, L., Yuan, J., Zhang, H., Jiang, X., Ma, J.: Piafusion: A progressive infrared and visible image fusion network based on illumination aware. Information Fusion  \textbf{83},  79--92 (2022)

\bibitem{tang2023datfuse}
Tang, W., He, F., Liu, Y., Duan, Y., Si, T.: Datfuse: Infrared and visible image fusion via dual attention transformer. IEEE Transactions on Circuits and Systems for Video Technology  (2023)

\bibitem{toet2014}
Toet, A.: Tno image fusion dataset. \url{https://doi.org/10.6084/m9.figshare.1008029.v2} (2014), figshare. Dataset

\bibitem{wang2021dual}
Wang, J., Liu, A., Yin, Z., Liu, S., Tang, S., Liu, X.: Dual attention suppression attack: Generate adversarial camouflage in physical world. In: Proceedings of the IEEE/CVF Conference on Computer Vision and Pattern Recognition. pp. 8565--8574 (2021)

\bibitem{XIE2023101835}
Xie, H., Zhang, Y., Qiu, J., Zhai, X., Liu, X., Yang, Y., Zhao, S., Luo, Y., Zhong, J.: Semantics lead all: Towards unified image registration and fusion from a semantic perspective. Information Fusion  \textbf{98},  101835 (2023)

\bibitem{xu2020u2fusion}
Xu, H., Ma, J., Jiang, J., Guo, X., Ling, H.: U2fusion: A unified unsupervised image fusion network. IEEE Transactions on Pattern Analysis and Machine Intelligence  \textbf{44}(1),  502--518 (2020)

\bibitem{yue2023dif}
Yue, J., Fang, L., Xia, S., Deng, Y., Ma, J.: Dif-fusion: Towards high color fidelity in infrared and visible image fusion with diffusion models. IEEE Transactions on Image Processing  (2023)

\bibitem{zhang2021image}
Zhang, H., Xu, H., Tian, X., Jiang, J., Ma, J.: Image fusion meets deep learning: A survey and perspective. Information Fusion  \textbf{76},  323--336 (2021)

\bibitem{zhang2018unreasonable}
Zhang, R., Isola, P., Efros, A.A., Shechtman, E., Wang, O.: The unreasonable effectiveness of deep features as a perceptual metric. In: Proceedings of the IEEE conference on computer vision and pattern recognition. pp. 586--595 (2018)

\bibitem{zhao2023ddfm}
Zhao, Z., Bai, H., Zhu, Y., Zhang, J., Xu, S., Zhang, Y., Zhang, K., Meng, D., Timofte, R., Van~Gool, L.: Ddfm: denoising diffusion model for multi-modality image fusion. In: Proceedings of the IEEE/CVF International Conference on Computer Vision. pp. 8082--8093 (2023)

\bibitem{zhu2017fusion}
Zhu, P., Ma, X., Huang, Z.: Fusion of infrared-visible images using improved multi-scale top-hat transform and suitable fusion rules. Infrared Physics \& Technology  \textbf{81},  282--295 (2017)

\end{thebibliography}

\end{document}